\documentclass[conference]{IEEEtran}
\usepackage{times}

\usepackage[numbers]{natbib}
\usepackage{multicol}
\usepackage[bookmarks=true]{hyperref}

\usepackage{graphics} 
\usepackage{epsfig} 
\usepackage{amsmath} 
\usepackage[section]{placeins}
\usepackage[linesnumbered, ruled]{algorithm2e}
\usepackage{multirow}
\usepackage{xcolor}
\usepackage{subcaption}
\usepackage[utf8]{inputenc}
\usepackage{amsmath}

\SetCommentSty{mycommfont}

\begin{document}

\title{ Pose Imitation Constraints for Collaborative Robots}

\author{Glebys Gonzalez$^{1}$ and Juan Wachs$^{2}$
\thanks{*This work was not supported by any organization}
\thanks{$^{1}$Glebys Gonzalez is a Research Assistant at the
Department of Industrial Engineering,
        Purdue university, 350 N Street, West Lafayette, IN 47096, USA
        {\tt\small gonza337@purdue.edu}}%
\thanks{$^{2}$Juan Wachs is an Associate Professor at the
Department of Industrial Engineering,
        Purdue university, 350 N Street, West Lafayette, IN 47096, USA
        {\tt\small jpwachs@purdue.edu}}%
}



%

\maketitle

\begin{abstract}
Achieving human-like motion in robots has been a fundamental goal 
in many areas of robotics research. 
Inverse kinematic (IK) solvers have been
explored as a solution to provide kinematic structures with
anthropomorphic movements. In particular, numeric solvers based on
geometry, such as FABRIK, have shown potential for producing human-like
motion at a low computational cost. Nevertheless, these methods  have shown limitations when solving for robot kinematic constraints. This work proposes a framework inspired by FABRIK for human pose imitation in real-time. The goal is to mitigate the problems of the original algorithm while retaining the resulting human-like fluidity and low cost. We first propose a human constraint model for pose imitation. Then, we present  a pose imitation algorithm (PIC), and it's soft version (PICs) that can successfully imitate human
poses using the proposed constraint system. PIC was tested on two collaborative robots
(Baxter and YuMi). Fifty human demonstrations were collected for a bi-manual assembly and an incision task. Then, two performance metrics were obtained for both robots: pose accuracy with respect to the human and the percentage of environment occlusion/obstruction. The performance of PIC and PICs was compared against the numerical solver baseline (FABRIK). 
The proposed
algorithms achieve a higher pose accuracy than FABRIK for both tasks (25\%-\textit{FABRIK}, 53\%-\textit{PICs}, 58\%-\textit{PICs}). 
In addition, PIC and it's soft version achieve a lower percentage of occlusion during 
incision (10\%-\textit{FABRIK}, 4\%-\textit{PICs}, 9\%-\textit{PICs}).
These results indicate that the PIC method can reproduce human poses and achieve key desired effects of human imitation.

\end{abstract}

\IEEEpeerreviewmaketitle

\section{Introduction}

Endowing robots with human-like motion has been a long
desired goal for roboticists, operators and casual users, specially in  
areas such as human robot collaboration \cite{Do2008ImitationOptimization,Hussein2017ImitationLearning,Maeda2017ProbabilisticTasks,Zheng2015ImpactsHandovers}, 
social robotics \cite{Zaraki2017DesignInteraction}
and imitation learning \cite{Tai2016AImitation,Argall2009ADemonstration}.
Inverse Kinematic (IK) solvers have been used to convert human arms' exhibited trajectories to robot joint motions and configurations that resemble the human movement. 
Nevertheless, current IK methods do not provide "solutions" that resemble the human motion when the DOF of the manipulators differ significantly from those of the human arm \cite{Aristidou2018InverseSurvey} leading to resulting poses that look unnatural, or singularities along the
trajectory causing instability. Such motions are unnatural in the sense that they lack motion economy and self-occlusion of the working area or operating field, as opposed to human motion during work \cite{Hussein2017ImitationLearning,Zheng2015ImpactsHandovers}. 
Aristidou et. al proposed a IK
method called FABRIK \cite{Aristidou2011FABRIK:Problem,Aristidou2016ExtendingConstraints}
that showed potential for tackling these problems. FABRIK produces smooth, human-like motions at a very low computational cost in human avatars and complex worm-type IK-structures. Nevertheless, when this algorithm is applied to manipulators, because their joint configuration differs from the humans', the solver can produce less natural poses and is often unable to converge to a solution \cite{Aristidou2018InverseSurvey}.  Thus, this paper proposes a novel constraint model for human pose and motion
imitation and IK solver for Pose Imitation
Constraints (PIC) inspired by the FABRIK algorithm. PIC aims to produce natural
human-like motions of the original algorithm, while mitigating the
problems that appear when applying FABRIK to manipulators when resembling human motion

First, we present a system of robot constraints for human pose imitation.
In our algorithm we define a model for the human pose and then map that model to the robot.
To achieve this human-robot mapping, we mark an arbitrary subset of three joints 
in the manipulator (sequentially selected) as the
'shoulder', 'elbow' and 'wrist'. In this model, the constraints 
are not represented as rigid angles between joints 
\cite{Alibeigi2017InverseTechnology} 
but as solution spaces for the selected 'shoulder', 'elbow'
and 'wrist' links.  
Then, the pose model is solved by PIC. This algorithm borrows the iterative
backward-forward approach to converge to a pose from FABRIK, but adds a different
constraint treatment. In FABRIK, every joint restriction is assumed to be independent
from the other joints. Conversely, in PIC, the backward loop is used to adjust for
the constraints that depend on previous links. This allows to optimize for the human constraint model at very low computational cost (at most 20 iterations). In addition,
we propose a soft version of the PIC
algorithm called PICs. The PICs method expands the pose constraint ranges
of the robot joints when a solution cannot be found under the original setup of constraints.

Human pose imitation by robots is desirable because it makes the 
tasks performed by robots more legible and transparent to humans,
which improves human-robot collaboration 
\cite{Hussein2017ImitationLearning,Zheng2015ImpactsHandovers}.
Thus, we chose two tasks for human pose imitation and
measured pose similarity and target occlusion.
In the assembly task, we focus on two actions: in the first, one agent aligns two blocks that will need to be assembled together in the next turn by a collaborator. The second task is a
incision on a suture kit, where one agent follows the incision lines on the pad demonstrated by another agent. 

Two robots with very different kinematic structures attempted to imitate the movements shown by the human:
a simplified Baxter Robot and an ABB YuMi 1400. Three different algorithms
were tested on the imitation phase: (1) The proposed PIC, (2) the
soft constrain version of the method PICs and (3) the original FABRIK \cite{Aristidou2011FABRIK:Problem}.
This work shows that both PIC and its soft version PICs perform better than FABRIK 
in pose similarity and target occlusion/obstruction, while maintaining the low computational
complexity and the motion smoothness of the original algorithm.
On average, PICs increased
the pose similarity achieved through FABRIK by 56\%. In addition, PICs  reduced the self occlusion during incision by 60\% and PIC reduced the obstruction during assembly by 22\%.

This work has three main contributions: 
1) A pose imitation constraint model that easily maps humans movements to robots naturally, 
2) Two flavors of a pose imitation solver (PIC and PICs)
that can be generalized to a variety kinematic
structures and 3) A case study involving a simulation of two different robots
imitating human task performance during different tasks in real time.

\section{Previous work}

Creating solvers that can imitate human motions
has been thoroughly discussed in \cite{Aristidou2018InverseSurvey}.
Consequently, we divide solver techniques into three main groups: analytical, numerical and heuristic.
Most of the methods in the analytical category determine
closed form solutions that can map human motion to manipulator's kinematics. 
The work in \cite{RileyEnablingHumanoids} uses analytic expressions
to map the human's position and orientation to a humanoid's reference frame. Then, the IK problem is solved for each joint sequentially and later concatenating those partial solutions into a single one. In \cite{Kulpa2005FastFigures}
analytical IK expressions were determined for each human joint and then an iterative solver
was applied to enforce kinematic and balance constraints in a robot. More recently,
the work in \cite{Alibeigi2017InverseTechnology} proposed closed form formulas to map
 human joint angles to a NAO robot's joint angles.
 These constrains were projected onto the null space of the Jacobian to produce anthropomorphic motions. Analytical methods have the disadvantage of producing 
 singularities when following human poses and trajectories. In addition, these analytical solvers depend on the task work-space and robot links' physical dimensions, making these methods difficult to generalize.  

Numerical methods for pose imitation as applied in a procedure as follows:
First, the observed human movements 
are scaled to the robot's workspace. Then, a numerical solution 
with additional constraints (in the robot's joint angles or velocities)
is used to replicate the demonstrated motions. 
The method in \cite{Kim2009StableMotions} presented a
dynamically stable imitation algorithm that uses scaling and
zero moment points to map the original trajectory to a more feasible one. Then, 
the new trajectory is solved using the IK method developed 
by \cite{Kim2005SolvingOptimization}. 
Another technique models the desired posture
as a set of target joint torques and a center of pressure \cite{Yamane2009SimultaneousData}.
The goal consists of minimizing the difference between the real and 
the target joint torques. In \cite{NakazawaImitatingAnalysis},
a method is proposed to imitate dancing motions by dividing the movement into motion
primitives and then using the solver proposed by \cite{PollardAdaptingRobot}
to adapt the joint angles and velocities to a humanoid robot. These works have two main 
disadvantages over the heuristic methods: they have a high computational cost and need 
fine tuning during the optimization steps \cite{Aristidou2018InverseSurvey}.

Heuristic based methods are iterative low-cost solvers that rely on simpler
formulations of the kinematic problem. The CCD method \cite{Wang1991AManipulators},
a popular heuristic solver, has been adapted and extended to produce
anthropomorphous motion, such as in \cite{Kulpa2005Morphology-independentAnimation}.
This method scales the original poses to fit the target structure.
The modified CCD works only if there is a resemblance between the human and
the target robot/mimicker. In addition, Kenwright
et al, fine-tuned the CCD algorithm to produce more anthropomorphous movements
\cite{Kenwright2012InverseCCD}. This implementation can use complex IK structures, unlike
the original method. Such approaches carry some of the weaknesses of the original CCD,
such as unstable solutions. Conversely, FABRIK is another solver that
has become popular in recent years due to it's human-like smooth performance
\cite{Aristidou2011FABRIK:Problem}. This method runs a "backward" and a
"forward" pass on a structure until the target movement is reached. 
FABRIK has been enhanced to incorporate human constraints \cite{Aristidou2016ExtendingConstraints}.
This constrained version of the algorithm has been used for imitating locomotion \cite{Agrawal2016Task-basedLocomotion} 
in simulated humans but has not shown equal success with manipulator movements \cite{Aristidou2018InverseSurvey}.
Our work proposes a formulation for human pose imitation that leverages
on the FABRIK algorithm\cite{Aristidou2011FABRIK:Problem}.
We propose a human constraint model that can 
be applied to a variety kinematic chains while keep a natural movement appearance and through economy of motion.


\section{Methods}
A pose can be defined by the relative position
between the joints. In the human arm there are three joints (excluding the hand): the shoulder,
the elbow, and the wrist. The 27 hand joints \cite{YingWu2001HandRecognition}
are not being considered in this paper, since most robotic 
grippers only have one to three DOF. Unlike joint based
constraints, the Pose Imitation Constraints (PIC) are designed to be 
independent of the morphology
of each robot or its parameters. 

\subsection{Anthropomorphous Constraints}
\begin{figure}[h!]
\centering
\begin{subfigure}[b]{0.45\textwidth}
   \includegraphics[width=1\linewidth]{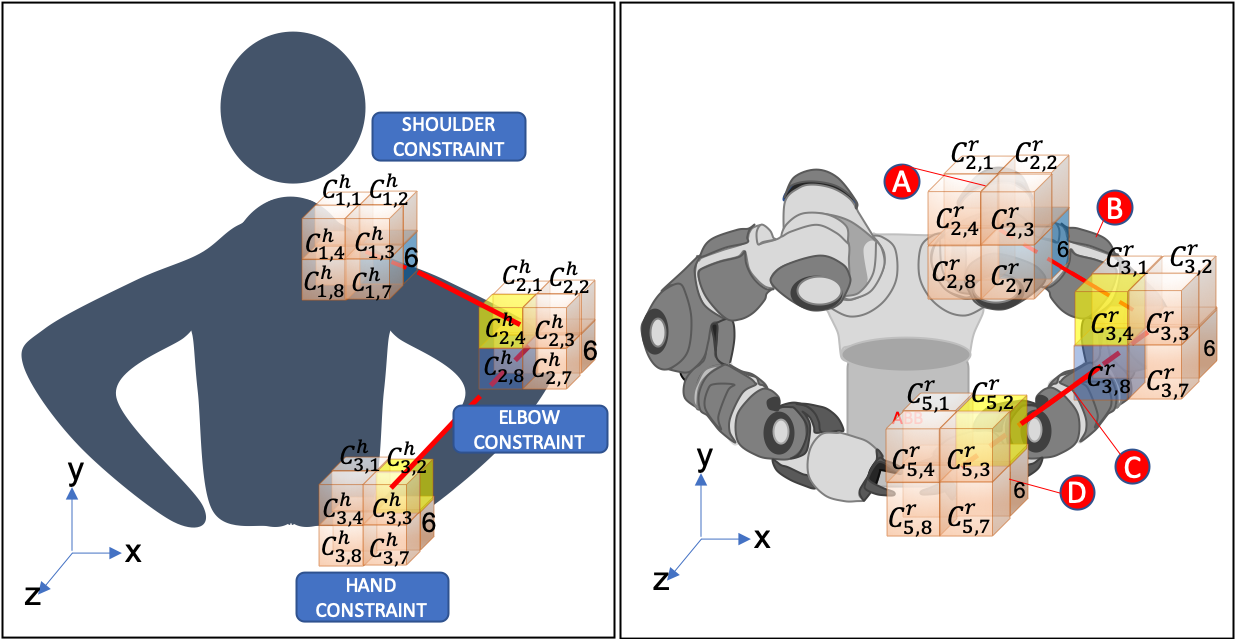}
   \caption{}
   \label{Figure:constraints_1} 
\end{subfigure}

\begin{subfigure}[b]{0.2\textwidth}
   \includegraphics[width=1\linewidth]{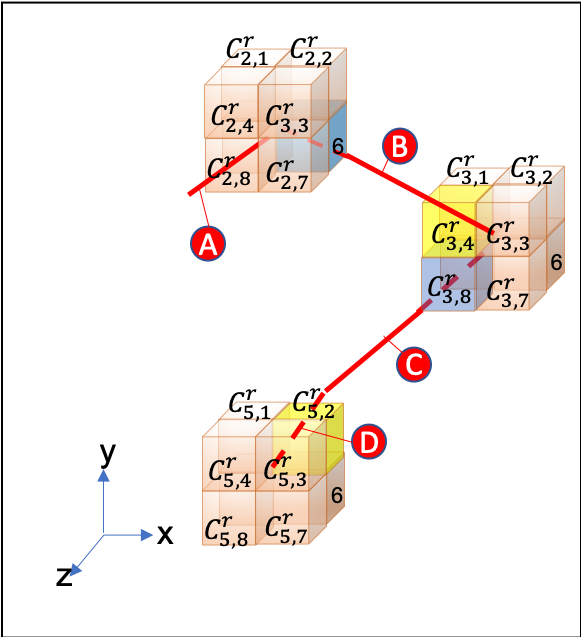}
   \caption{}
   \label{Figure:constraints_2}
\end{subfigure}
\caption{(a) Left: Human PIC constraints (a) Right: Mapping of the human PIC to the robot system. (b) detailed image of the constraint
system where the human shoulder, elbow and wrist are mapped to the robot joints 2, 3 and 5.}
\label{Figure:constraints_full}
\end{figure}

This section defines a set of concepts necessary to model a 
human pose. Let $H_i$, were $i \in \{1,\dots,3\}$ be the set of human 
joints in the human arm (shoulder, elbow and wrist). Let $C_i^h, i=1,\dots,3$
be a coordinate system located at joint $H_i$ aligned with the world's reference
frame (the $h$ denotes that the coordinate system is attached to a human joint). 
Each coordinate frame $C_i^h$ is divided into 8 octants, where $c_{i,k}^h$ is 
the $k\textsuperscript{th}$ octant of the $C_i^h$ coordinate frame 
(see Figure \ref{Figure:constraints_1}). Octants are the 
extension of quadrants to three dimensions.
Finally, a human pose constraint is defined by a pair of octants
$(c_{i,k}^h,c_{i+1,m}^h)$, where 
$i \in \{1, 2\}$ and $k,m \in \{1,\dots,8\}$. 
The first octant $c_{i,k}^h$ defines the constraint
space for the link attached to joint $i$. We define this octant area
as an \textit{OUT} constraint, since the link going \textit{OUT} of joint $i$
has a range of motion limited to octant $c_{i,k}$. The next octant ($c_{i+i,m}^h$)
is defined as an \textit{IN} constraint, since it defines the allowed area for
 the previous link attached to joint $i+1$. The example in Figure \ref{Figure:constraints_1} shows the \textit{OUT} constraints in blue and the \textit{IN} constraints 
 in yellow.

A human  pose $P$ will be defined by two pairs of constraints.
One between the shoulder and elbow and one between the elbow and wrist
(See Equation \ref{eq:constraints}). For example, the constraint model shown 
in Figure \ref{Figure:constraints_1} (Left), can be represented by the following octant pairs: 
$P_h=[(c_{1,6}^h,c_{2,4}^h),(c_{2,8}^h,c_{3,2}^h)]$.
Without loss of generality, for any two joints, and a
division into $O$ octants we get:

\begin{multline}
    P_h = [(c_{i,k}^h,c_{i+i,m}^h),(c_{i+i,l^h},c_{i+2,n}^h)], \\
    \textrm{ where } k,m,l,n \in \{1,\dots,O\} 
\label{eq:constraints}
\end{multline}

The purpose of this constraint system is
to define a model in a space that can be easily mapped to 
a variety of robots. This has the potential of facilitating imitation tasks in classrooms with heterogeneous robots.
Let $J = \{ J_1, J_2, \dots, J_N\}$ be the set of joints of a robot, where $N$
is the total number of joints in the system. Similar to the human model, we 
create a set of  3 coordinate frames that are aligned with the robot's wold 
reference frame. Each reference frame is then defined as: $C_i^r, i=1,\dots,N$
where $i \in \{1,\dots,N\}$ and the origin of the frame is located at joint $J_i$ (the $r$ denotes the coordinate system as attached to a robot joint).
These coordinate systems are also divided into octant regions, where $c_{i,k}^r$ is 
the $k\textsuperscript{th}$ octant of the $C_i^r$ coordinate frame. Finally, we define
a robot pose $P_r$ using 4 octants as we did with the human pose: 
$P_r=[(c_{i,k}^r,c_{i+i,m}^r),(c_{i+i,l^r},c_{i+2,n}^r)]$, where  
$k,m,l,n \in \{1,\dots,O\}$. 

Let $\Phi: P_h \longrightarrow P_r$ be the function that maps the constraint 
system from the human to the robot (with each $C_i^r$ is divided into $O$ octants). The mapping $\Phi$ assigns the three human joints
to any three robot joints and then applies the 
octant constraints at those respective joints.
Equation \ref{eq:mapping} defines $\Phi$ given that the human
shoulder, elbow and wrist are mapped
to the robot joints $r_s,r_e$ and $r_w$. Figure \ref{Figure:constraints_1} shows an
example mapping from human to robot constraints. Figure \ref{Figure:constraints_2}
shows the resulting mapping in more detail.

\begin{multline}
    \Phi(P_h, r_s,r_e,r_w)= P_r \\
    \Phi([(c_{i,k}^h,c_{i+i,m}^h),(c_{i+i,l^h},c_{i+2,n}^h)], r_s,r_e,r_w) =   \\
    [(c_{r_e,k}^r,c_{r_s,m}^r),(c_{r_s,l}^r,c_{r_w,n}^r)], \\
    \textrm{ where } k,m,l,n \in \{1,\dots,O\} 
    \label{eq:mapping}
\end{multline}

\subsection{PIC and PICs algorithms}
In the previous section we introduced two types of constraints: \textit{IN} and \textit{OUT}.
An \textit{OUT} constraint can be treated as a regular constraint on the range of movement of a joint. Conversely, the \textit{IN} constraint determines the position where the end of the link is located and therefore restricts the region at which
that link can connect with the next one. In this section we will explain how the PIC and PICs
algorithms can optimize for both constraint types.

\subsubsection{FABRIK review}
The PIC algorithm is based on the FABRIK method \cite{Zhang2018FABRIKc:Robotsb}. FABRIK
proposes to reach a target by performing a "backward" step followed by a "forward" step in a loop.
The "backward" pass places the final link at the target we want to reach and tilt this link
towards the end of the previous link. The same process is done to all the subsequent links in decreasing
order. By the end of the "backward" pass the IK structure has
a joint configuration that is similar to the original one, with the gripper at the target point and the first link displaced from the base. Then, a "forward" pass is performed, placing the first link at the base and tilting it towards the second link. This step is repeated for all
the links in increasing order. When the "forward" pass is over, the IK structure displays a pose much 
closer to the goal than it was in the previous step. The backward and forward steps are performed iteratively until a minimum error between the current and target position is reached.

\begin{figure}[htb!]
\centering
\begin{subfigure}[b]{0.95\columnwidth}
   \includegraphics[width=1\linewidth]{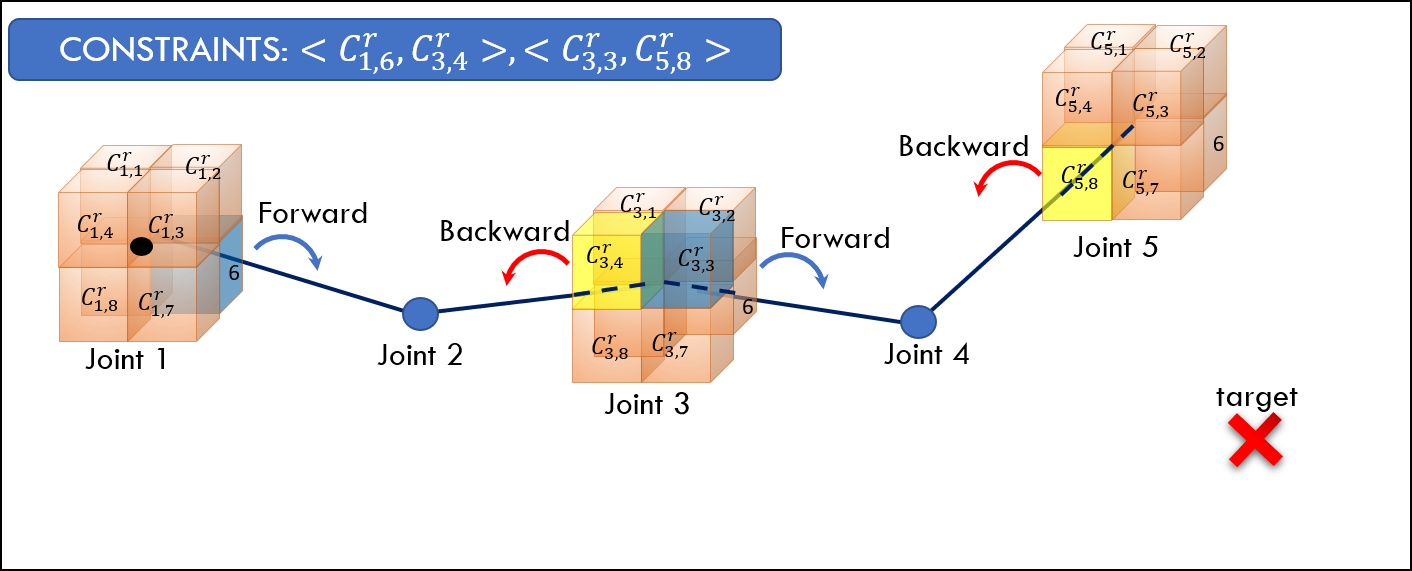}
   \caption{}
   \label{Figure:PIC_1} 
\end{subfigure}
\begin{subfigure}[b]{0.95\columnwidth}
   \includegraphics[width=1\linewidth]{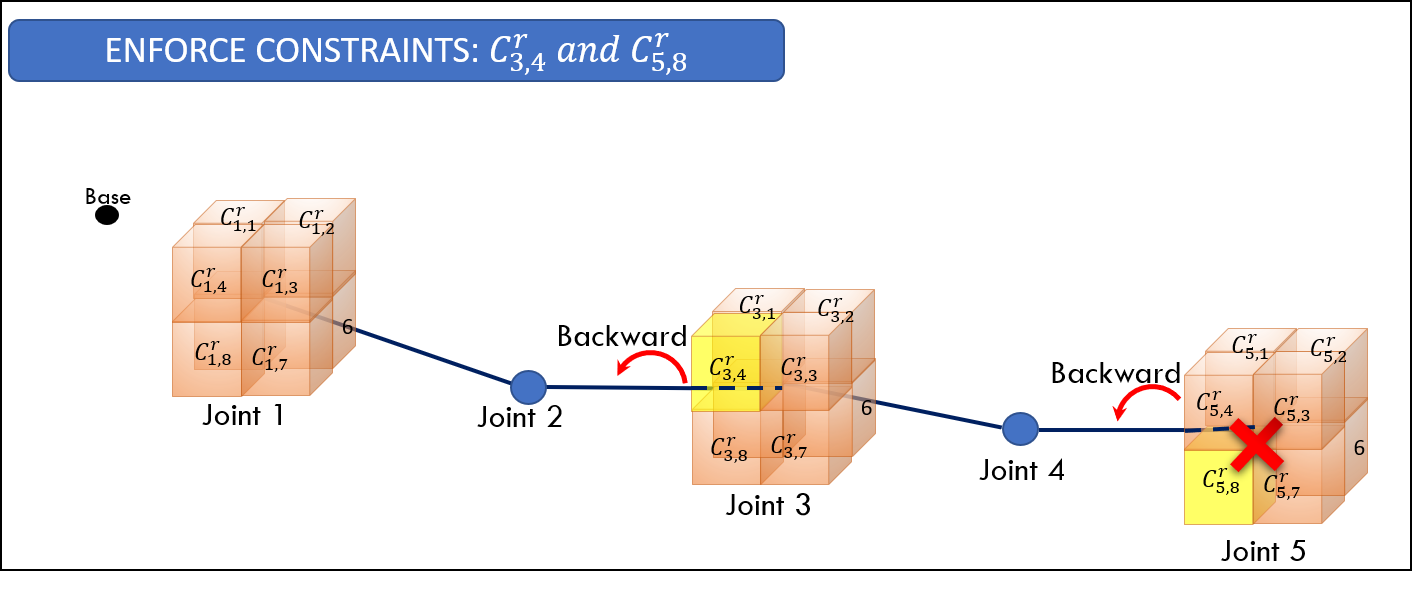}
   \caption{}
   \label{Figure:PIC_2}
\end{subfigure}
\begin{subfigure}[b]{0.95\columnwidth}
   \includegraphics[width=1\linewidth]{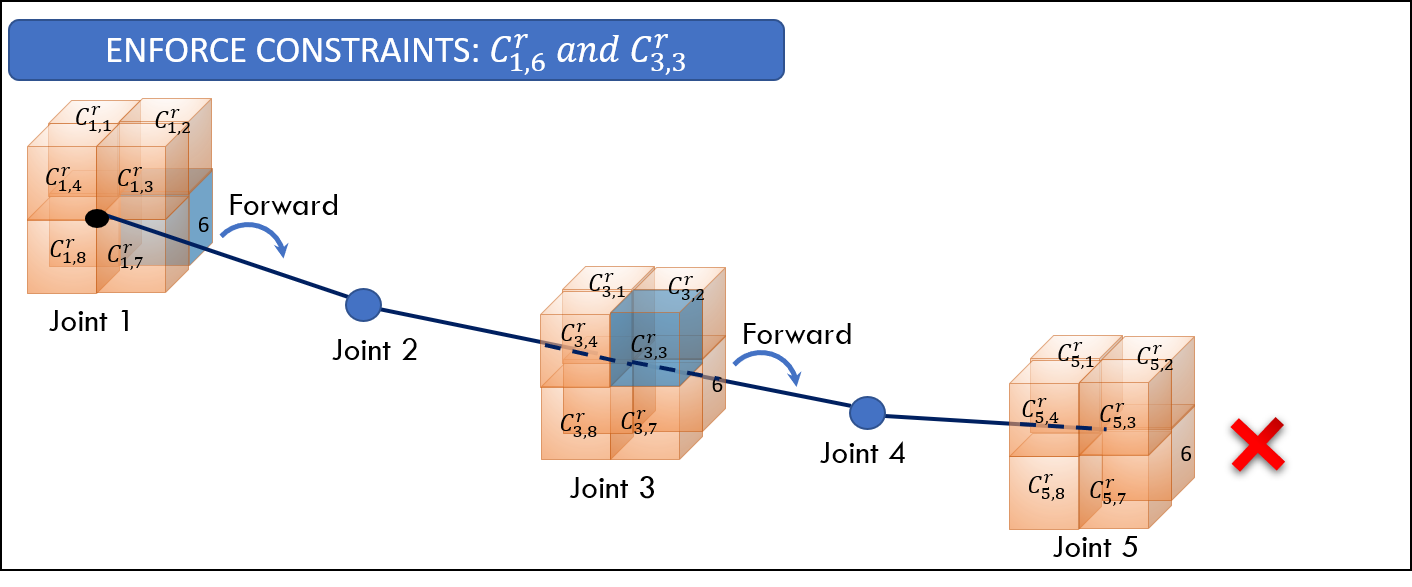}
   \caption{}
   \label{Figure:PIC_3}
\end{subfigure}
\caption{(a) Summary of the PIC iteration with an example constraint configuration. (b) Forward step of the PIC loop. (c) Backward step of the PIC loop}
\label{Figure:PIC_full}
\end{figure}

\subsubsection{PIC algorithm}
The proposed PIC algorithm optimizes the  \textit{IN} constraints in the backward pass and
the \textit{OUT} constraints in the forward pass, as shown in Figure \ref{Figure:PIC_1}.
The \textit{OUT} constraints (see Figure \ref{Figure:PIC_3}) are easy to handle because they depend
solely on the joint where the constraint is located. Since the forward pass determines the final
orientation of each link, the PIC algorithm just needs to 
restrict the range of motion of the links to the specified octants. 

The \textit{IN} constraint is more complex to enforce because it depends on the solutions
of the previous joints. For example, in Figure \ref{Figure:PIC_2} the constraint $C_{3,4}$
depends on the solutions for joins 1 and 2 and the constraint $C_{5_7}$ depends on the
joints 1 to 4. The backward loop determines the joint positions and orientations in a reverse
order (from the gripper to the base). Creating a constraint at the joint $i$ would force
the previous $i-1$ joints to adjust to the orientation change. Thus, the PIC enforces the 
\textit{IN} constraints at the backward pass. At the the beginning of the PIC run, 
the \textit{IN} constraints are set and updated with the forward pass, but as the algorithm converges, the changes in the
kinematic chain become smaller and the constraints in the backward loop are preserved. 

\begin{figure}[!htb]
    \centering
    \includegraphics[width=0.8\linewidth]{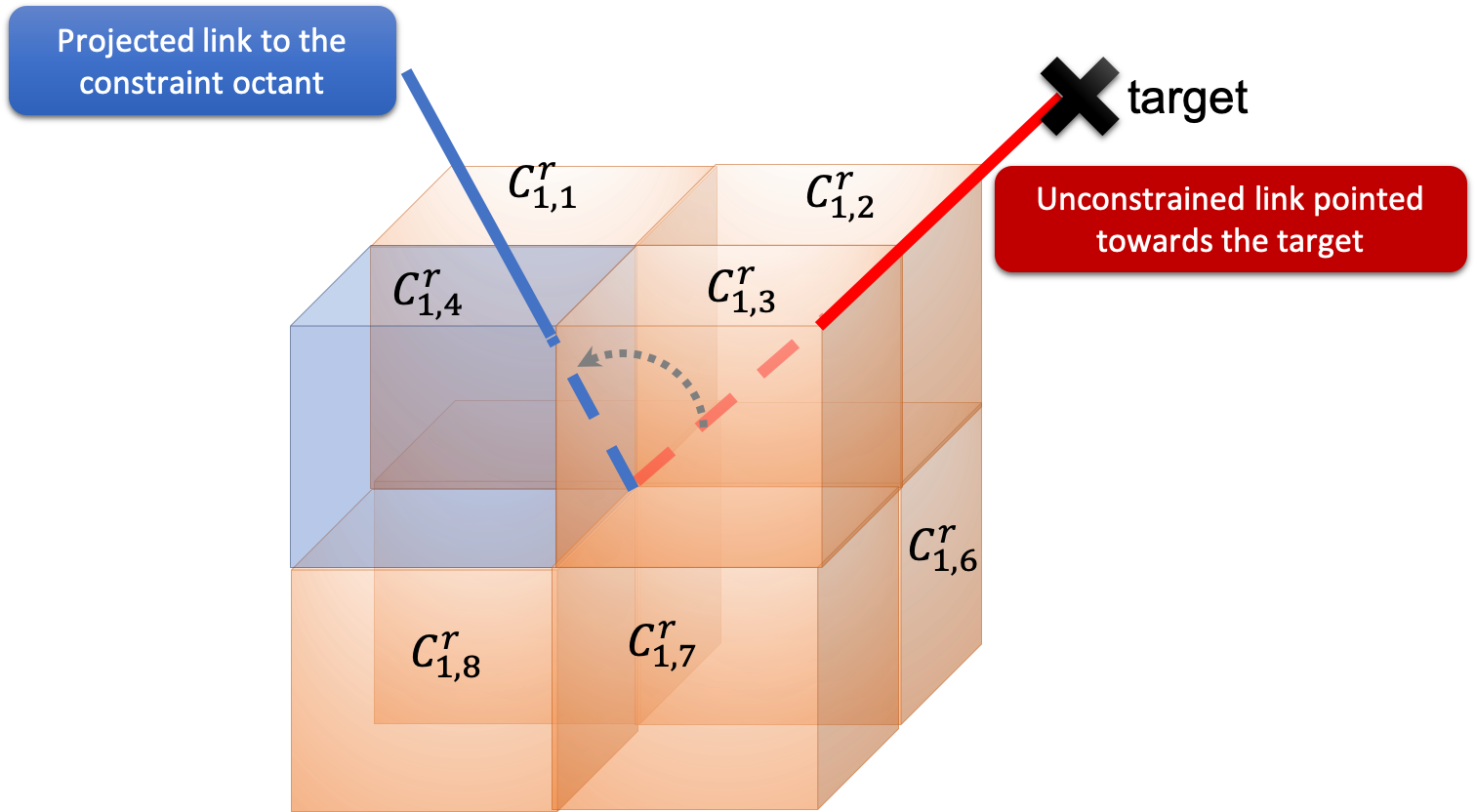}
    \caption{Example of the constraint checking step. 
    When the unconstrained link (shown in red) falls outside of the
    allowed Octant (i.e $C^r_{1,4}$), the algorithm projects the link to the
    closest face/plane of that Octant.}
    \label{Figure:octact_project}
\end{figure}

Finally, if orientation of a link does not fall inside the Octant region, PIC projects the constrained link to the closest plane to that Octant, as shown in Figure \ref{Figure:octact_project}.

\subsubsection{PICs algorithm}
Constraining the robot to maintain the pose defined by a mapped $P_r$
logically decreases the size of its solution space.
This might cause the target pose to be unreachable by the robot. In order
to alleviate this problem, we developed a constraint softening method called Pose Imitation Constraint-soft (PICs).
The PICs algorithm uses the same backward-forward logic to optimize for \textit{IN}-\textit{OUT} constraints.
The difference lies on the link orientation calculation: when a link
falls outside of the constraint region,
PICs adds the the neighboring octants to the admissible range of motion,
allowing the link to get closer to it's intended target.
Finally, a neighboring Octant can be defined as follows:
Let $k$ and $q$ be the $k^{th}$ and $q^{th}$ Octants in a coordinate 
system $C$ at any joint. 
Let $axis_k = (x_k, y_k, z_k)$ and $axis_q= (x_q, y_q, z_q)$ 
be the sings of the $x,y,z$ axis for $k$ and $q$ at the coordinate system $C$. 
We say $k$ is a neighbor of $q$ if the following condition holds:

\begin{equation}
    Hamming(axis_k, axis_1) <= \eta
\label{eq:neighbors}
\end{equation}

Where $Hammimg()$ is the hamming distance between sequences \cite{Bookstein2002GeneralizedDistance} and $\eta$ is the softening factor.
The softening factor $\eta$ is the neighbor distance allowed by the
constraints. If $\eta =1$, the neighboring Octants are allowed to have only one axis that does not match. Thus, Octants with two common axis can be added to the range of motion. Subsequently, if $\eta =2$ the Octants with one common axis can be added to the link's range of motion.

\section{Experiments}
Fifty human/teacher demonstrations were collected to be imitated by a simplified 
YuMi and a Baxter robot. 
The tasks were recorded with a Kinect 2 and the skeleton data obtained by the 
Kinect sensor for each arm were 
used as inputs for this experiment. First, the position signals of each 
joint  filtered independently using exponential smoothing. Then, the filtered signals obtained from the
human skeleton were multiplied by the a transformation matrix, 
so the captured motion would match the robot motion's position and scale. 

\subsection{Tasks}
\begin{figure}[h]
    \centering
    \begin{subfigure}[b]{0.75\columnwidth}
        \includegraphics[width=\textwidth]{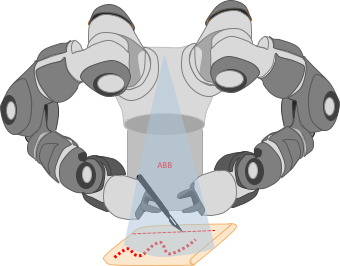}
        \caption{Experimental setup for incision task}
        \label{Figure:setup_1}
    \end{subfigure}
    ~ 
    \begin{subfigure}[b]{0.95\columnwidth}
        \includegraphics[width=\textwidth]{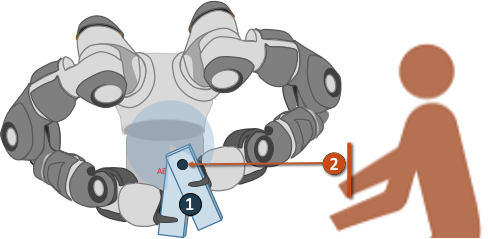}
        \caption{Experimental setup for Assembly task}
        \label{Figure:setup_2}
    \end{subfigure}
    \caption{Experimental setup for the selected tasks}
    \label{Figure:setup}
\end{figure}

Two tasks were proposed for this work. The first one in a surgical 
incision, with two possible variations: a) a straight line, b) a curve.
The second one is an assembly task, where two pieces had to be aligned 
with respect to a particular plane. This task was subdivided in three types,
depending on the plane that the human was aligning to. 
The tasks were recorded to study
the effect of the pose imitation with multiple robots.
The setting is that of a teacher in a classroom with two robot learners.
Figure  \ref{Figure:setup} shows the tasks in more detail.
For each task, ten different human samples were collected.
During the incision, it is desired for the robot to 
minimize the self occlusion, since surgeons do the same to
accurately evaluate their work.
For the assembly task, we assume there is a collaborator at the other side of the table. This collaborator would
continue with the next assembly step when the robot finishes aligning the pieces (see Figure \ref{Figure:setup_2}).
In the case of assembly is desirable to
minimize the robot obstruction, so the collaborator 
can easily perform the next step.

\subsection{Algorithms and metrics}
Three algorithms were implemented to perform the tasks:
1) PIC, 2) PICs and 3) FABRIK \cite{Aristidou2011FABRIK:Problem}.
To assess the pose imitation performance in multiple robots, a set of 
metrics were defined. First, we define the pose similarity according to \cite{HaradaQuantitativePerception} as the angle between the shoulder
and the wrist link. Then, we define the Pose Accuracy ($Pacc$) as the
normalized number of data points where mean squared error of two pose 
angles is less than a threshold $\delta$:

\begin{equation}
Pacc =\frac{1}{n} \Big(  
\sum_{i=1}^{n} Pacc_i \Big), \ 
Pacc_i
\left\{
	\begin{array}{ll}
		1  & \mbox{if } {(\theta_i^h - \theta_i^r)}^2 <\delta \\
		0 & \mbox{otherwise} \\ 
	\end{array}
\right.
\end{equation}

Where $\theta_i^h$ is the angle between the 
shoulder and elbow link for the human at a given frame $i$ where $i=1,\dots,N$.
The robot has an equivalent $\theta_i^r$ measurement. 

\begin{figure}
    \centering
    \begin{subfigure}[b]{0.45\columnwidth}
        \includegraphics[width=\textwidth]{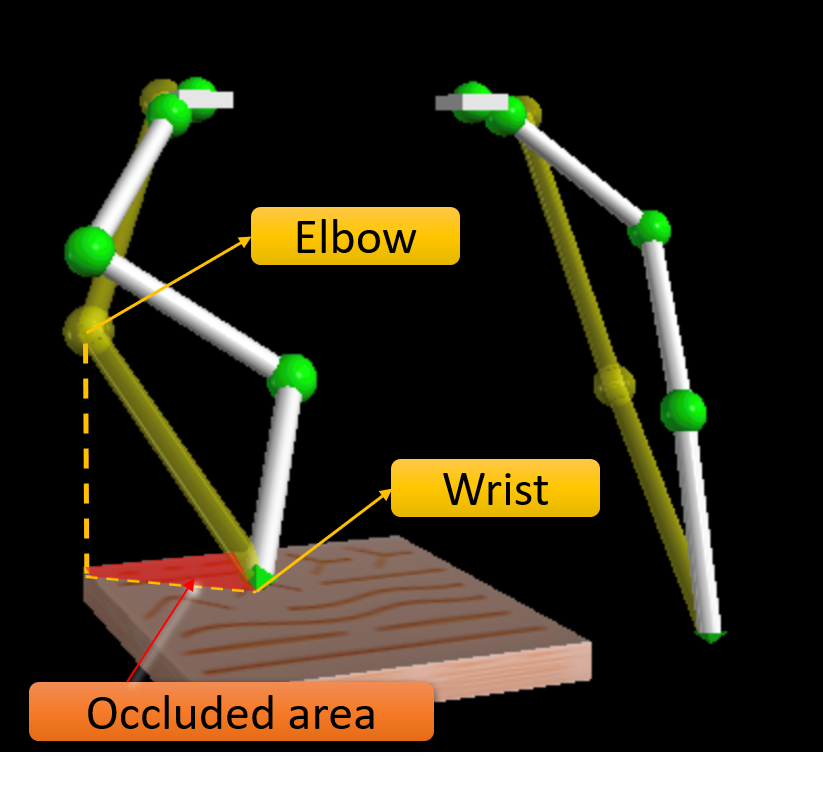}
        \caption{Human Occlusion}
        \label{Figure:visibility_1}
    \end{subfigure}
    ~ 
    \begin{subfigure}[b]{0.45\columnwidth}
        \includegraphics[width=\textwidth]{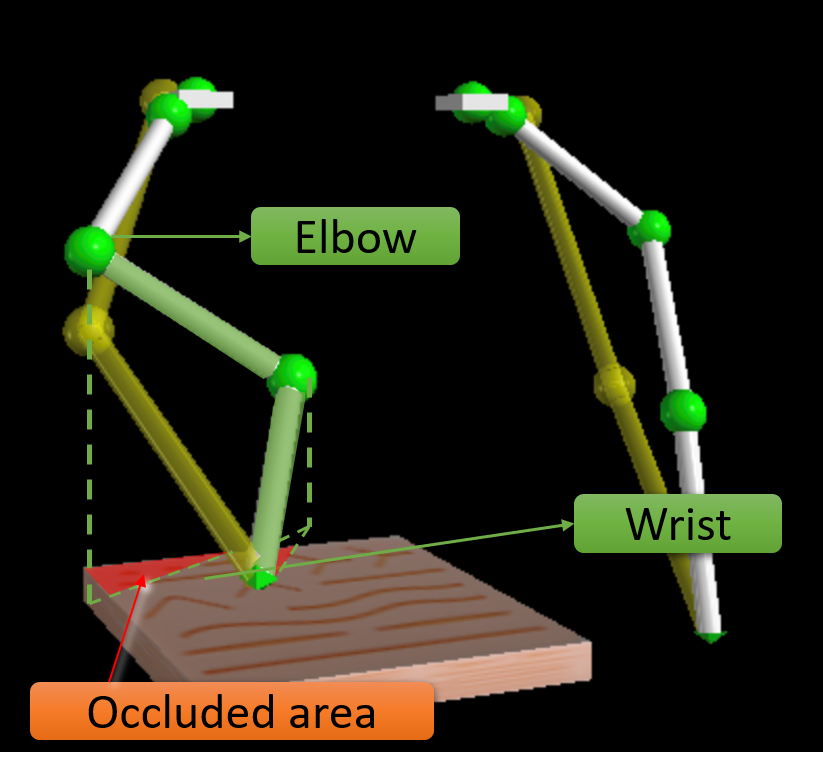}
        \caption{Robot Occlusion}
        \label{Figure:visibility_2}
    \end{subfigure}
    \caption{Left: Pad area occluded by the human. Right: Pad area occluded by the Robot}
    \label{Figure:visibility_all}
\end{figure}

A second metric mas created to measure the effect of PIC in the range of movement available for the robot and the collaborator. 
Let $PO$ be the the Percentage of Occlusion/Obstruction during a task. 
The $PO$ metric is based on the
occlusion that the robot/human creates on a Region of Interest (ROI) inside a plane. For the the incision task, we are interested on the 
occlusion caused to the plane and ROI that match the incision pad. Thus,
the $PO$ metric is obtained by projecting each link on
the plane that aligns with the surgical pad. Then,  
the area under each projected link is calculated and normalized by the ROI's area (See Figure \ref{Figure:visibility_all}).

Let $f_i$ be the  2D line equation that represents the projection of each 
link segment into the pad and $ROI_{area}$ be the area of the Region of Interest
in the plane. Then, the Percentage of Occlusion (PO) can be calculated as:

\begin{multline} \label{eq:po}
PO = \sum_i \int_{x_i}^{x_{i+i}}{ \frac{f_i}{ROI_{area}},} \\
\mbox{where }  f_i=
\left\{
	\begin{array}{ll}
		0 & \mbox{if } m_ix+b_i \leq 0 \\
		h & m_ix+b_i \geq h \\
		m_ix+b_i  & \mbox{otherwise}  \\
	\end{array}
\right.
\end{multline}

Where $x_i$ and $x_{i+1}$ are the values with respect to the  $X$-axis of the ROI's plane,  for link $i$ and link $i+1$ respectively, and  $h$ is the height of the ROI. Figure \ref{Figure:visibility_1}
illustrates the occluded area of the human skeleton (skeleton in yellow) and 
Figure \ref{Figure:visibility_2} shows the occluded area by the robot 
(robot skeleton in white and green). 

For the assembly task, the $PO$ metric was used to determine the percentage of 
obstruction for the collaborator.  In this scenario, the plane of projection was aligned with the insertion area (See blue region in Figure \ref{Figure:setup_2}). 
And the ROI's size was empirically determined to match the robot's size. 

\begin{table}[h]
\caption{Pose Accuracy with respect to the human motion}
\begin{center}
\begin{tabular}{|c|c||c|c|c|}
\hline
Robot &Task &FABRIK & PIC & PICs ($\eta=3$)  \\
\hline
\hline
\multirow{2}{*}{YuMi} & Incision - S & 0.17 & \textbf{0.84} & 0.67 \\
& Incision - C & 0.00 & \textbf{0.70} & 0.50\\
\hline
\multirow{2}{*}{Baxter} & Incision - S & 0.29 & \textbf{0.70} & 0.65\\
& Incision - C & 0.25 & \textbf{0.74} & 0.82 \\
\hline
\hline
\multirow{3}{*}{YuMi} & Assembly 1& 0.53 & 0.53& \textbf{0.68} \\
& Assembly 2 & 0.34 & 0.13 & \textbf{0.58} \\
& Assembly 3 & 0.52 & 0.57  & \textbf{0.81} \\
\hline
\multirow{3}{*}{Baxter} & Assembly 1& 0.00 & 0.09 & \textbf{0.43} \\
& Assembly 2 & 0.03 & \textbf{0.66} & 0.40 \\
& Assembly 3 & 0.31 & \textbf{0.32} & 0.21 \\
\hline
\multicolumn{2}{|c||}{\textit{\textbf{Method mean}}} & 0.25 & 0.53 &\textbf{ 0.58 } \\
\hline
\end{tabular}
\end{center}
\label{table:acc_results}
\end{table}

\subsection{RESULTS}
A Baxter and a YuMi robot performed 50 human demonstrations, as described
in the previous section. Three algorithms were evaluated: 1) The proposed
PIC, 2) PICs and 3) FABRIK \cite{Aristidou2011FABRIK:Problem}. 
The Pose Accuracy (Pacc) achieved for the incision and assembly tasks is shown in Table \ref{table:acc_results}. 

Both versions of the pose imitation algorithm  outperform FABRIK in average Pose Accuracy.
During the incision task, PIC achieved an average Pose Accuracy (Pacc) of 75\% ($\pm$16), PICs  ($\eta=3$) achieved a Pacc of 66\% ($\pm$16) while FABRIK produced a Pacc of 18\% ($\pm$14). In addition, both PIC and PICs outperformed FABRIK during the assembly task, obtaining an average Pose Accuracy of 38\%($\pm$31) and 52\%($\pm$29) 
respectively. In contrast, FABRIK only produced a pose similarity of 29\%($\pm$30) for this task. 
The assembly task proved to be much more challenging for 
both robots when it came to following the human pose, producing a higher variance. 
Finally, the two flavors of PIC showed a better Pose Accuracy than FABRIK 
for both Baxter and YuMi in all the task repetitions. This results indicate that the 
proposed algorithm can handle human imitation for kinematic structures that vary in size,
and DoF.

\begin{table}[h]
\caption{Percentage of Occlusion during Incision}
\begin{center}
\begin{tabular}{|@{\hskip2pt}c|@{\hskip1pt}c||@{\hskip2pt}c|@{\hskip2pt}c|@{\hskip2pt}c|@{\hskip2pt}c|}
\hline
Robot &Task &FABRIK & PIC & PICs ($\eta=3$)  & Human \\
\hline
\multirow{2}{*}{YuMi} & I-S & 0.10 & 0.02 & \textbf{0.01}  & 0.11\\
& I-C & 0.07 & \textbf{0.05 }& 0.07  & 0.09\\
\hline
\multirow{2}{*}{Baxter} & I-S & 0.11& \textbf{0.03}& 0.13  & 0.07\\
& I-C & 0.12 & \textbf{0.06} & 0.14  & 0.08 \\
\hline
\multicolumn{2}{|@{\hskip2pt}c}{\textit{\textbf{Method mean}}} & 0.10 & \textbf{0.04} & 0.09   & 0.09\\
\hline
\end{tabular}
\end{center}
\label{table:vis_results_incision}
\end{table}

\begin{table}[h]
\caption{Percentage of Obstruction during Assembly}
\begin{center}
\begin{tabular}{|@{\hskip2pt}c|@{\hskip1pt}c||@{\hskip2pt}c|@{\hskip2pt}c|@{\hskip2pt}c|@{\hskip2pt}c|}
\hline
Robot &Task &FABRIK & PIC & PICs ($\eta=3$) & Human \\
\hline
\multirow{3}{*}{YuMi} & A-1 &0.06 & 0.04 & \textbf{0.03}  & 0.06\\
& A-2 & 0.08 & 0.08 & \textbf{0.07} & 0.08 \\
& A-3 & \textbf{0.03} & 0.03 & \textbf{0.03}  & 0.03 \\
\hline
\multirow{3}{*}{Baxter} & A-1 & 0.08  & \textbf{0.05} & 0.07  & 0.06 \\
& A-2 & 0.20 & 0.18 & \textbf{0.15}  & 0.13\\
& A-3 & 0.12  & 0.08 &\textbf{0.07}  & 0.04 \\
\hline
\multicolumn{2}{|c}{\textit{\textbf{Method mean}}} & 0.09 & 0.08 & \textbf{0.07}  & 0.07\\
\hline
\end{tabular}
\end{center}
\label{table:vis_results_assembly}
\end{table}

\begin{figure}
    \centering
    \begin{subfigure}[b]{0.95\columnwidth}
        \includegraphics[width=\textwidth]{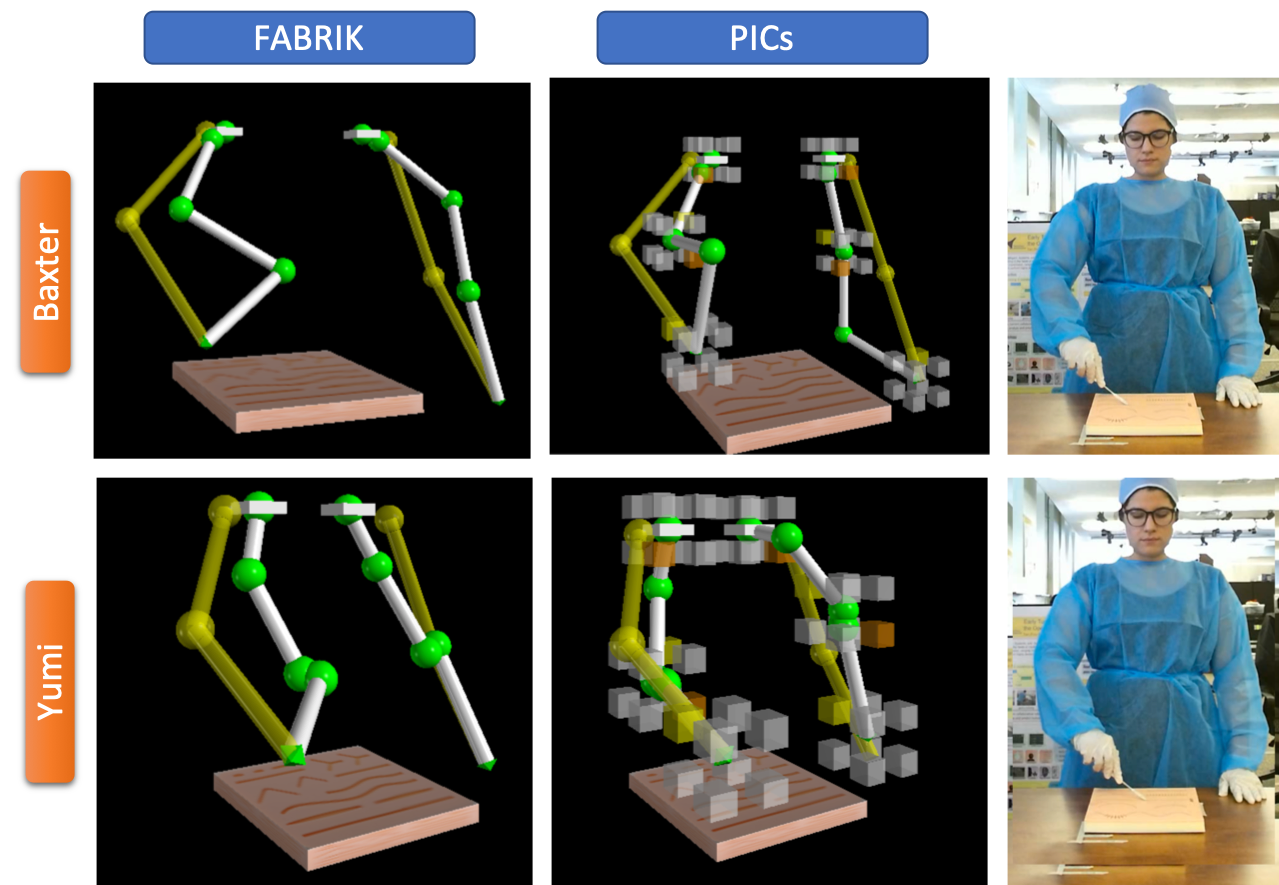}
        \caption{Pose examples for the incision task}
        \label{Figure:P_example_1}
    \end{subfigure}
    ~ 
    \begin{subfigure}[b]{0.95\columnwidth}
        \includegraphics[width=\textwidth]{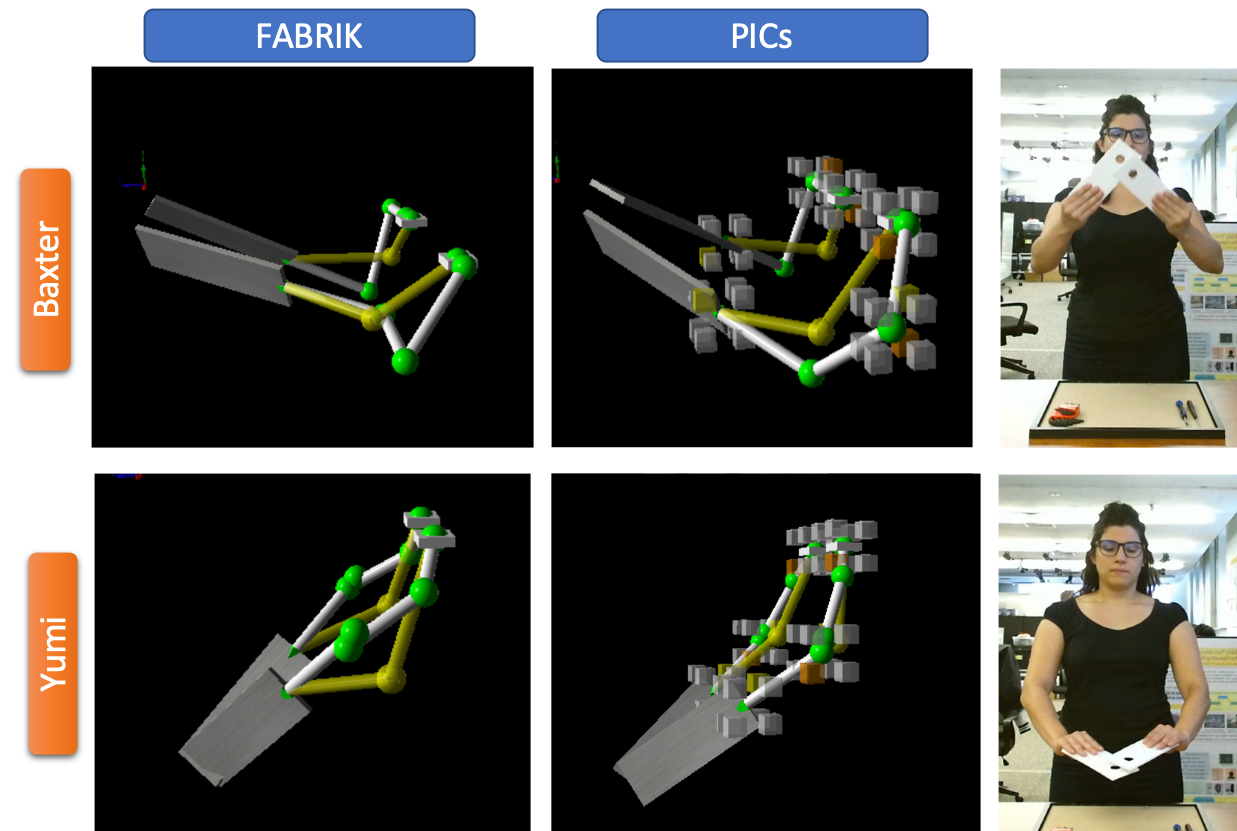}
        \caption{Pose examples for the assembly task}
        \label{Figure:P_example_2}
    \end{subfigure}
    \caption{PICs($\eta=3$) vs FABRIC resulting  pose examples. (a) Results during the incision. (b) Results during the assembly task}
    \label{Figure:P_example_all}
\end{figure}

The softening effect of the  PICs algorithm over the Pose Accuracy was calculated and compared against PIC and FABRIK.
Figure \ref{Figure:P_example_all} shows the average pose similarity
of both FABRIK and
PICs for all softening levels
$\eta=1,\dots,3$, over the two incisions and the three assemblies.  The results show that the higher the value of $\eta$ (the softening parameter),
the greater the average Pose Accuracy between both tasks.

This effect over the Pose Accuracy (Pacc) is likely due to the fact that in the PIC algorithm the \textit{OUT} constraints take precedence over 
the \textit{IN} constraints. This precedence occurs because the \textit{OUT} constraints are always enforced last. 
If the \textit{OUT} region is too limited, the algorithm might override the solution
for the \textit{IN} constraint found in the previous step.
Thus, a higher value for $\eta$ in PICs translates a to better pose imitation performance, 
because the algorithm allows a higher range of motion on each constrained joint.
This creates a bigger solution space where 
the \textit{OUT} constraints don't have to override the \textit{IN} constraints to find the target pose.

The Percentage of Occlusion/Obstruction (PO) was obtained for both incision and assembly. 
In each data sample, this metric was calculated only when
when the human wrist was hovering over the ROI (see Equation \ref{eq:po}).
Table \ref{table:vis_results_incision} shows the Percentage of Occlusion (self-occlusion) for the incision task. 
The FABRIK baseline achieves an average PO
of 10\% ($\pm$4\%) and its Pose Imitation counterparts
achieve an occlusion of 4\% ($\pm$3\%) and 9\% ($\pm$10\%) for PIC and PICs respectively.
Thus, PIC produces an Occlusion reduction of 56\% with respect to FABRIK.

\begin{figure}[h]
    \centering
    \includegraphics[width=0.95\linewidth]{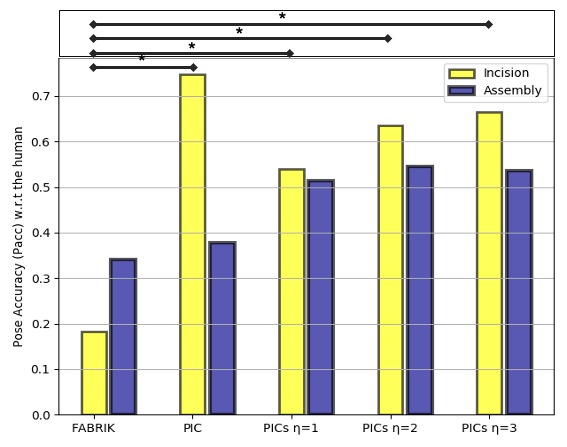}
    \caption{Average Pose Accuracy (Pacc) at different levels of softening, where
    $\eta=0$  refers to PIC with no softening. (*=statistical significance).}
    \label{Figure:poses}
\end{figure} 

In addition, the Percentage of Obstruction (PO) was studied 
for the assembly tasks (see Table \ref{table:vis_results_assembly}).
In this scenarios, FABRIC, PIC and PICs produced comparable results. In particular, FABRIC produced a PO of 9\%($\pm$6\%) PIC of 8\%($\pm$5\%) and PICs achieved 7\%($\pm$4\%) of obstruction. Nevertheless, the PICs method still reduced the collaborator obstruction  by 22\%.

For both tasks, PIC or it's soft version exhibited a smaller PO than FABRIK.  These results indicate that both pose imitation algorithms are better at 
minimizing the occlusion. In addition, 
the PIC algorithm produced an occlusion level that is very 
close to the original human's 
(see Tables \ref{table:vis_results_incision} and \ref{table:vis_results_assembly}.
As discussed above, the PIC shows a stronger preference for the
\textit{OUT} constraints. Thus, these constraints have a direct effect
on the occlusion.

\begin{figure}[h]
    \centering
    \includegraphics[width=0.95\linewidth]{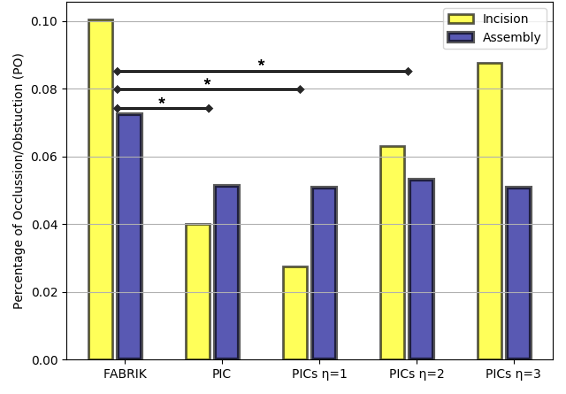}
    \caption{Percentage of Occlusion/Obstruction (PO) 
    at different levels of softening, where
    $\eta=0$  refers to PIC with no softening. (*=statistical significance).}
    \label{Figure:PO}
\end{figure}

The softening effect using the PICs algorithm was also studied for occlusion.
Figure \ref{Figure:PO} depicts the PO for FABRIK, PIC and 
for all levels of $\eta=1,\dots,3$ in PICs. 
Both PIC and PICs produce a Percentage of Occlusion significantly smaller than FABRIK.
The PICs method (using $\eta=1$) produces the smallest  PO, by only Occluding/Obstructing 5\% of the Region of Interest. 

The level of softening is a parameter that can be
tuned to maximize Pose Accuracy and and minimize self-occlusion. The aggregate results of incision and assembly show that PICs with $\eta=3$ produces the maximum combined average between
the two metrics: a Pose Accuracy (Pacc) of 58\% and a Percentage of Occlusion (PO) of 8\%. 

The current algorithm faces a problem that should be addressed in future work.
When the pose configuration changes, the PIC might not find a smooth solution
between the old and new pose constraints, this makes the robot jump to 
the new pose position. This problem can be solved by using a temporal window
of $N$ before and after steps. The window would allow to discern if the current
data point should be eliminated or if an interpolation should be added to create a smoother transition. \cite{GonzalezPIC:Algorithm.}.

\section{Conclusion} 
\label{sec:conclusion}

This work introduces a framework for specifying human constraints for pose imitation during task performance.
Such framework constitutes the basis of an imitation algorithm (Pose Imitation Constraints - PIC), based on
the FABRIK method \cite{Aristidou2011FABRIK:Problem}. The PIC algorithm
 can successfully reproduce the poses of a human teacher in multiple robots.
 This method was tested during a two different tasks, an assembly and an incision,  using two simulated robots (ReThink Baxter and a ABB YuMi). 
 In addition, a variation to the PIC using soft constraints (PICs) is proposed. 
 The results show that the PICs algorithm outperforms FABRIK in pose similarity: 66\% vs 18\% for incision and 52\% vs 29\% during assembly. In particular, PICs increased the overall Pose Accuracy
 obtained for FABRIK by 56\%. Both PIC and PICs produce less occlusions during the task. 
 Particularly, the PIC achieves a percentage of self-occlusion
of 4\%, while FABRIK achieves 10\% during the incision task. 
The proposed imitation strategy can be used as the first step in a robot coaching process
\cite{Riley2006Coaching:Behaviors} as part of a major vision involving human teaching heterogeneous robots in classrooms.
\cite{Petric2014OnlineGestures,Calinon2007ActiveDemonstration,Lee2011IncrementalTube}.
The proposed algorithm
could help improve the quality of robot imitation performance and reduce the number of 
iterations that it takes to teach a task with lower errors and wider operational space available to the collaborator/teacher. 



\bibliographystyle{plainnat}
\bibliography{paper_template}

\end{document}